\documentclass{article} 
\usepackage{nips13submit_e,times}
\usepackage{amsmath,amssymb,graphicx,enumitem,breqn,wrapfig}
\usepackage[space]{grffile}
\usepackage[hang,raggedright]{subfigure}
\usepackage{hyperref}
\usepackage{url}

\title{Learning Features and their Transformations by Spatial and Temporal Spherical Clustering}
\author{Jayanta K. Dutta and Bonny Banerjee\\
Institute for Intelligent Systems, and Department of Electrical \& Computer Emgineering\\
University of Memphis, Memphis, TN 38152, USA\\
\emph{jkdutta@memphis.edu}~,~\emph{bonnybanerjee@yahoo.com}}

\nipsfinalcopy 

\begin{document}

\maketitle

\begin{abstract}
Learning features invariant to arbitrary transformations in the data is a requirement for any recognition system, biological or artificial. It is now widely accepted that simple cells in the primary visual cortex respond to features while the complex cells respond to features invariant to different transformations. We present a novel two-layered feedforward neural model that learns features in the first layer by spatial spherical clustering and invariance to transformations in the second layer by temporal spherical clustering. Learning occurs in an online and unsupervised manner following the Hebbian rule. When exposed to natural videos acquired by a camera mounted on a cat's head, the first and second layer neurons in our model develop simple and complex cell-like receptive field properties. The model can predict by learning lateral connections among the first layer neurons. A topographic map to their spatial features emerges by exponentially decaying the flow of activation with distance from one neuron to another in the first layer that fire in close temporal proximity, thereby minimizing the pooling length in an online manner simultaneously with feature learning.
\end{abstract}

\section{Introduction}
\label{Sec:Introduction}

Learning features invariant to arbitrary transformations in data is a requirement for any biological or artificial recognition system. In recent years, there has been a surge of interest in learning feature hierarchies from data using multilayered or deep learning models largely motivated by the layered organization of the neocortex. It is now widely accepted that the simple and complex layers in the primary visual cortex (or V1) are responsible for learning transformation-invariant features \cite{HubelWiesel1962}. A number of computational models have been proposed that can learn transformation-invariant features for state-of-the-art recognition in images, audio and videos using alternating simple and complex layers, such as Neocognitron \cite{Fukushima2003}, convolutional neural networks \cite{LeCunBengio1995,Ciresanetal2012} and HMAX \cite{RiesenhuberPoggio1999,Serreetal2007recognition}. Given the biological relevance and technological usefulness of the simple-complex layers, it is imperative to understand their function as a canonical computational unit.

The contribution of this paper is a fully-learnable model with only two manually tunable parameters, the learning rate and threshold decay parameter, for learning invariant features from spatiotemporal data, such that the model may be used as a canonical computational unit in deep neural networks for recognition and prediction. In the proposed model, the functions of the simple and complex layers have similar formulation in space and time respectively signifying their functional similarity.

In particular, we present a two-layered neural model (architecture and learning algorithm) that operates in a feedforward (or bottom-up), unsupervised and online manner. We show that:\\
1. Spatial features may be learned in the first layer by spatial clustering on the surface of a hypersphere of unit radius (a.k.a. \emph{spherical clustering} \cite{DhillonModha2001}) where the outliers are not allowed to influence the cluster centers. When learned from natural videos, these features resemble the receptive fields (RFs) of simple cells in V1. We will refer to the first layer of our model as the simple layer.\\
2. Arbitrary transformations of these features may be learned from time-varying data in the second layer by temporal spherical clustering where the outliers are not allowed to influence the cluster centers. When learned from natural videos, the response properties of the second layer neurons resemble that of complex cells in V1. We will refer to the second layer as the complex layer.\\
3. Predictive capability may be induced in this model by learning transition probabilities in lateral connections among the simple layer neurons. Higher-order predictions may be made by using the transformations learned in the complex layer in conjunction with the lateral connections.\\
4. A topographic map of the spatial features emerges by exponentially decaying the flow of activation with distance from one neuron to another in the same layer that fire in close temporal proximity. Unlike other models (e.g., \cite{HyvarinenHoyer2001,Kavukcuogluetal2009}) where the pooling regions are predefined or some sort of group sparsity is assumed to learn topographic maps from spatial data, we exploit the temporal continuity of data and physical constraints to learn topographic feature map.

\subsection{Receptive fields}
\label{Sec:Receptive fields}

In our model, the goal of simple and complex neurons may be conceptualized as learning subsets from a pool of neurons in space and time respectively. A postsynaptic neuron integrates activations from presynaptic neurons over space and time. Each neuron has a spatial RF and a temporal RF, both of fixed sizes. The size of a feature it will encode may be less than or equal to its spatial and temporal RF sizes. All neurons in a layer have the same sized RFs. The size of spatial RF of a simple neuron in layer $\l$ is defined by the number of neurons in the lower layer (i.e., layer $\l-1$) reporting to it at any time instant. The goal of the simple neuron in layer $\l$ is to get strongly connected (connections may be excitatory or inhibitory) to a subset of neurons within its spatial RF in order to encode the recurring spatially coincident patterns in layer $\l-1$. When this subset of presynaptic neurons fire, the postsynaptic simple neuron is highly likely to fire. The spatial RF size of a complex neuron is unity, i.e. at any time instant, a postsynaptic complex neuron can receive input from only one presynaptic simple neuron. This implicitly assumes the winner-take-all mechanism in the simple layer. All simple neurons in layer $\l$ report to all complex neurons in layer $\l+1$ albeit at different time instants.

In our model, a neuron in layer $\l$ samples the input stream every $\tau^{(\l)}$ instants of time, where $\tau^{(\l)}$ is referred to as the temporal RF size of the neuron. Complex neurons in layer $\l+1$ sample the input at a lower frequency than simple neurons in layer $\l$, i.e. $\tau^{(\l+1)} > \tau^{(\l)}$. Conceptually, a neuron in layer $\l$ fails to distinguish the temporal sequence of events occurring within $\tau^{(\l)}$ instants of time, and hence considers all of those events to occur simultaneously. However, a neuron in layer $\l-1$ can keep track of the temporal sequence of events occurring within $\tau^{(\l)}$ time instants due to higher sampling frequency. We use this insight to model the feedforward weights to learn sets and the lateral weights in conjunction with feedforward weights to learn sequences.

The goal of a complex neuron in layer $\l$ is to get strongly connected to a subset of simple neurons in layer $\l-1$ in order to encode the recurring temporally coincident patterns in layer $\l-1$ where each pattern corresponds to an instance of an arbitrary transformation. The size of this subset is at most $\tau^{(\l)}$. When this subset of presynaptic simple neurons fire in close temporal proximity (i.e., within $\tau^{(\l)}$ time instants), the postsynaptic complex neuron is highly likely to fire. The temporal RF size of a simple neuron is unity with respect to that of neurons in the lower layer, i.e. a postsynaptic simple neuron integrates activations from its lower layer over only one time instant. Thus, a simple neuron integrates activations from presynaptic neurons over space and fires if its threshold is crossed while a complex neuron integrates activations from presynaptic simple neurons over time and fires if its threshold is crossed. Then, being able to learn sets from a pool of neurons in space and time is the crux of the invariant feature learning problem.

\subsection{Objective function}
\label{Sec:Objective function}

Formally, we define a set $\mathcal{X}$ as a finite collection of distinct alphabets, written as $\mathcal{X} = \{x_{1}, x_{2}, ... x_{N}\}$ where $x_{i}$ is a $d$-dimensional alphabet or feature or event, $i\neq j$ implies $x_{i}\neq x_{j}$, and $N$ is the cardinality of $\mathcal{X}$, i.e. $|\mathcal{X}|=N$. We define a sequence $\zeta$ over the set $\mathcal{X}$ as a finite ordered list of alphabets from $\mathcal{X}$, written as $\zeta = \langle x_{1}, x_{2}, ... x_{n}\rangle$ where $x_{i}\in \mathcal{X}$, $i<j$ implies $x_{i}$ occurs before $x_{j}$, $i\neq j$ does not imply $x_{i}\neq x_{j}$, and $n$ is the length of $\zeta$. Therefore, learning a subset of features from recurring coincidences in the data requires clustering $\mathcal{X}$ into a set of $k$ clusters $\mathcal{C}=\{\mathcal{C}_{1}, \mathcal{C}_{2}, ... \mathcal{C}_{k}\}$. Soft-clustering is a better option for natural data.

Formation of a cluster may be viewed as a pseudo-event that occurs \emph{where} (in case of spatial clustering) or \emph{when} (in case of temporal clustering) all or most of the events in the cluster occur. Let,

\begin{equation}
S^{(\l-1)}_{i}(p) =
  \left\{
  \begin{array}{ll}
    1, & \mbox{if $x_{i}$ occurs at $p$}
\\
    0, & \mbox{otherwise}
  \end{array}
  \right.
\end{equation}

\begin{equation}
S^{(\l)}_{j}(p) =
  \left\{
  \begin{array}{ll}
    1, & \mbox{if $\mathcal{C}_{j}$ occurs at $p$}
\\
    0, & \mbox{otherwise}
  \end{array}
  \right.
\end{equation}

\noindent where $p$ denotes location in case of spatial clustering and time in case of temporal. Also,

\begin{equation}
W^{(\l-1,\l)}_{ij} = Pr(S^{(\l)}_{j}=1~|~S^{(\l-1)}_{i}=1)
\end{equation}

\noindent Clustering may then be defined as an optimization problem that minimizes the following objective function:

\begin{equation}
\label{Equ:Objective function}
\ell(\mathcal{S}^{(\l-1)},\mathcal{S}^{(\l)};W^{(\l-1,\l)}) =\\ \frac{1}{2} \displaystyle\sum_{j=1}^{k}\sum_{i=1}^{N} \|W^{(\l-1,\l)}_{ij} - \frac{1}{|P^{(\l)}_{j}|}\sum_{p\in P^{(\l)}_{j}} S^{(\l-1)}_{i}(p)\|^{2}
\end{equation}

\noindent where $W^{(\l-1,\l)}=[W^{(\l-1,\l)}_{ij}]_{N\times k}$ are the parameters of the model, $\mathcal{S}^{(\l-1)}=[S^{(\l-1)}_{i}(p)]_{N\times \mathcal{P}}$ and $\mathcal{S}^{(\l)}=[S^{(\l)}_{j}(p)]_{k\times \mathcal{P}}$ ($p=1,...\mathcal{P}$) are the observations. $P^{(\l)}_{j}=\{p:S^{(\l)}_{j}(p)=1\}$. The $W^{(\l-1,\l)}$ that minimizes $\ell$ is a maximum a posteriori probability (MAP) estimate assuming uniform prior. This formulation is similar to \emph{correlation clustering} \cite{BagonGalun2011}; it automatically recovers the underlying number of clusters $k$.

In the next section, we present a two-layered neural network model where neurons in lower layer are activated by spatial pseudo-events in $\mathcal{C}$ while those in higher layer are activated by temporal pseudo-events in $\mathcal{C}$. The feedforward weights are learned using the simplest form of Hebbian rule to minimize $\ell$ in an unsupervised and online manner. In Section \ref{Sec:Experimental results}, we describe experimental results on natural spatiotemporal data followed by conclusions.

\section{Network model}
\label{Sec:Network model}

\subsection{Architecture}
\label{Sec:Architecture}

\begin{figure}[t]
\centering
    \subfigure[] 
    {
        \label{Fig:AllLayers}
        \includegraphics[width=0.3\textwidth, totalheight=0.24\textheight]{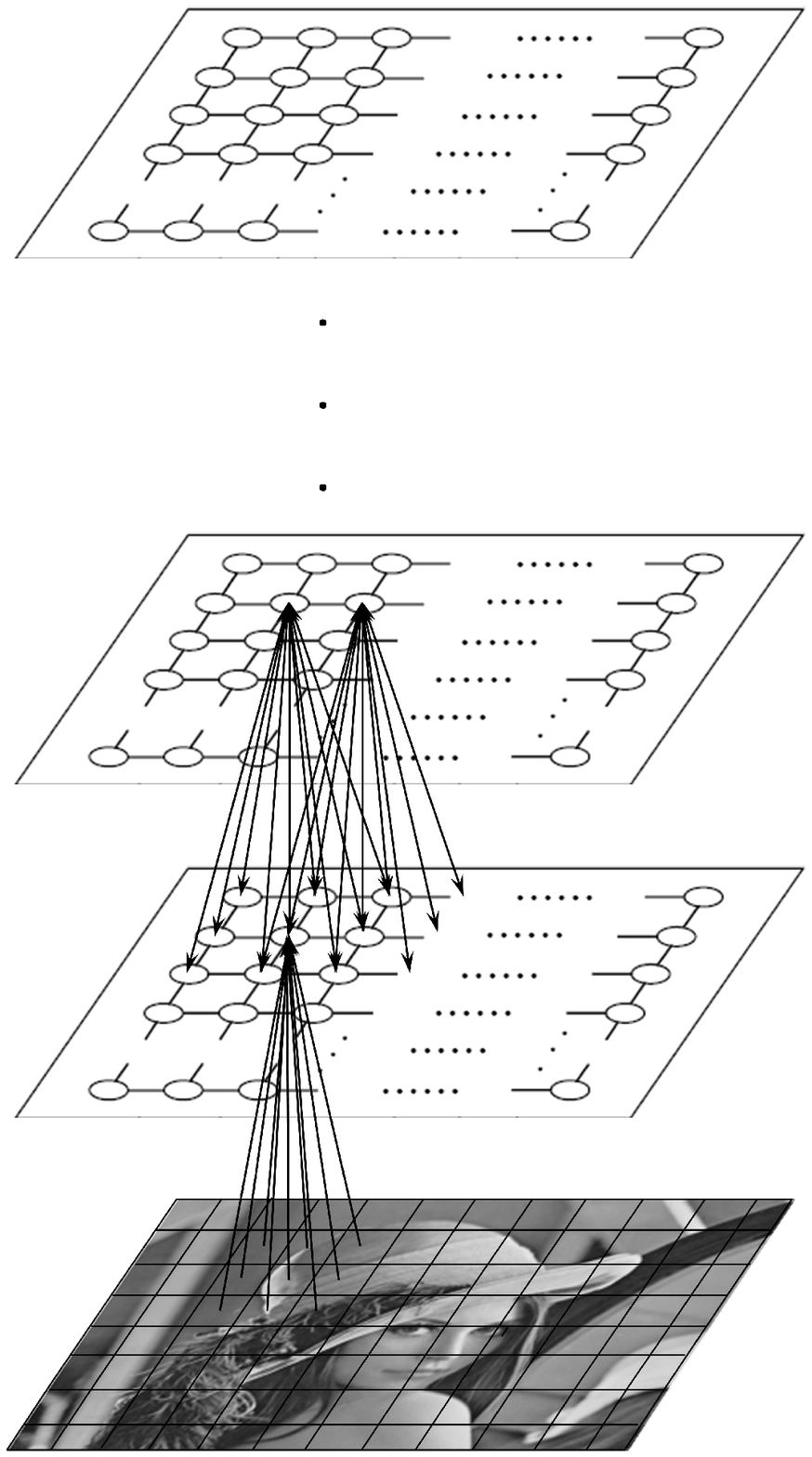}
    }
    \subfigure[] 
    {
        \label{Fig:FeedforwardConnections}
        \includegraphics[width=0.33\textwidth]{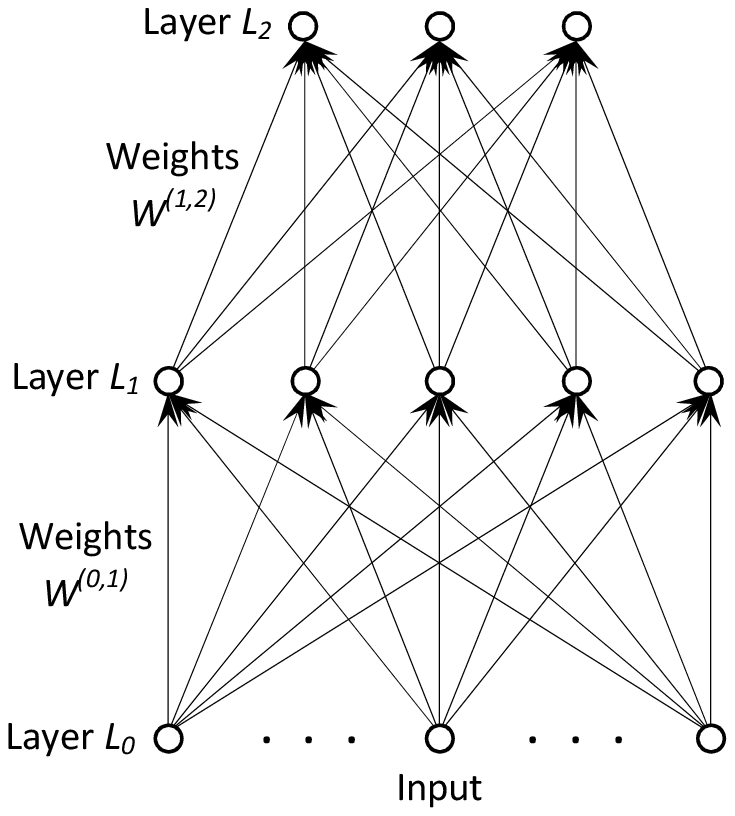}
    }
    \subfigure[] 
    {
        \label{Fig:LateralConnections}
        \includegraphics[width=0.3\textwidth]{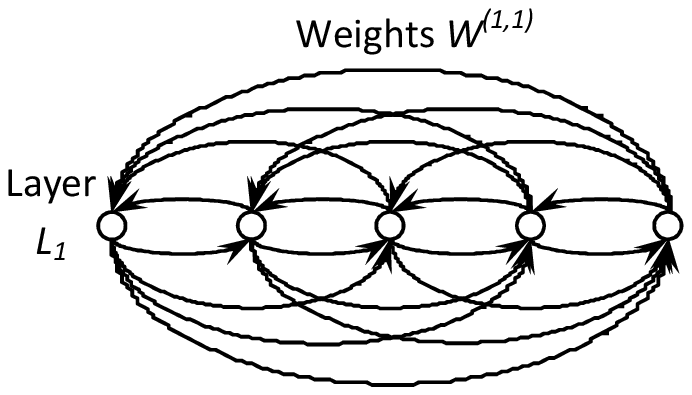}
    }
    \caption{(a) A hierarchical architecture is shown. Each layer denotes a pair of simple and complex layers. The circles denote nodes. (b) Feedforward connections from a simple layer ($L_{1}$) to a complex layer ($L_{2}$) within a node are shown. The circles denote neurons. The weights $W^{(0,1)}$ are learned to encode spatial sets or features in $L_{1}$. The weights $W^{(1,2)}$ are learned to encode temporal sets or transformations in $L_{2}$. (c) Lateral connections in a simple layer within a node are shown. The weights $W^{(1,1)}$ in conjunction with $W^{(1,2)}$ are modeled to learn sequences.}
\label{Fig:Architecture} 
\end{figure}

Our network architecture consists of a hierarchy of layers of nodes (see Fig. \ref{Fig:AllLayers}). A node is a canonical computational unit consisting of two layers -- simple neurons in the lower layer, and complex neurons in the higher layer (see Fig. \ref{Fig:FeedforwardConnections}). Neurons in a node are sparsely connected to neurons in the neighboring nodes in the same layer, one layer above and one layer below by lateral, feedforward and feedback connections respectively. The first (or lowest) layer in the hierarchy receives input from external data varying in space and time. In this paper, we will concentrate on learning invariant features \emph{in a node} using the feedforward and lateral connections only.

We will refer to the layer that receives external inputs as the input layer, denoted by $L_{0}$. The simple and complex layers in a node will be denoted by $L_{1}$ and $L_{2}$ respectively. Each neuron in $L_{1}$ is connected to all neurons in $L_{2}$ in a feedforward manner (see Fig. \ref{Fig:FeedforwardConnections}). The feedforward weights are denoted by $W^{(1,2)}$, where $W^{(k,\l)}_{ij}(t)$ is the weight or strength of connection from the $i^{th}$ neuron in layer $k$ to the $j^{th}$ neuron in layer $\l$ at time $t$. Each neuron in $L_{1}$ is also connected to all neurons in its own layer, except itself, by lateral connections (see Fig. \ref{Fig:LateralConnections}). The lateral weights are denoted by $W^{(1,1)}$. We will assume the number of neurons in $L_{0}$ reporting to the simple layer in a node is equal to the spatial RF size of a simple neuron in $L_{1}$ and that each neuron in $L_{0}$ is connected to all neurons in $L_{1}$ in a feedforward manner.

\subsection{Operation}
\label{Sec:Operation}

The goal of feedforward processing in the perceptual cortices is hypothesized to be rapid categorization \cite{Serreetal2007categorization}. Our model samples the input stream at regular intervals of time. At each sampling instant, it accepts spatial data as input through $L_{0}$ which is passed on to $L_{1}$ and $L_{2}$ in the form of activations. The goal of computations in a node is to selectively cluster the input into groups. Over time, each simple neuron in a node gets tuned to a unique feature which represents a spatial cluster center while each complex neuron gets tuned to a unique transformation. Functionally, a node is a bag of invariant filters all of which are applied to each patch of the input data.

\subsection{Neuron}
\label{Sec:Neuron}

A simple neuron in $L_{1}$ integrates activations from presynaptic neurons in $L_{0}$ over its spatial RF and fires if the integrated input crosses its threshold. Activations of simple neurons in $L_{1}$ at time $t$ are:

\begin{equation}
\label{Equ:SimpleActivation}
A^{(1)}(t) = A^{(0)}(t)\times W^{(0,1)}(t) + S^{(1)}(t-1)\times W^{(1,1)}(t)
\end{equation}

\noindent where $A^{(0)}$ is the external input, $S^{(1)}$ are the states of neurons in $L_{1}$, $A^{(0)}\times W^{(0,1)}$ and $S^{(1)}\times W^{(1,1)}$ are activations due to feedforward and lateral interactions respectively. Each feature in $W^{(0,1)}$, and $A^{(0)}$ are normalized, hence $A^{(1)}$ is the normalized dot product of the input with each feature modulated by the lateral interaction. This allows a simple neuron to act as a suspicious coincidence detector \cite{Foldiak1990}, responding with high activation if the input matches the feature encoded in its RF.

A complex neuron in $L_{2}$ integrates activations from presynaptic neurons in $L_{1}$ over its temporal RF and fires if the integrated input crosses its threshold. The activations of complex neurons in $L_{2}$ at time $t$ are:

\begin{equation}
\label{Equ:ComplexActivation}
A^{(2)}(t) = \displaystyle\sum_{h=t_{0}}^{t} S^{(1)}(h)\times W^{(1,2)}(h)
\end{equation}

\noindent where $t_{0}$ is a time instant from when the neurons start integrating, $t-t_{0}\leq \tau^{(2)}$. Each feature in $W^{(1,2)}$ is normalized to have unit norm. A complex neuron acts as a temporal coincidence detector.

The state of the $i^{th}$ neuron in any layer $\l$ is binary, given by

\begin{equation}
S^{(\l)}_{i}(t) =
  \left\{
  \begin{array}{ll}
    1, & \mbox{if $A^{(\l)}_{i}(t) > A^{(\l)}_{j}(t), \forall j\neq i$, and $A^{(\l)}_{i}(t) > \theta^{(\l)}_{i}(t)$}
\\
    0, & \mbox{otherwise}
  \end{array}
  \right.
\end{equation}

\noindent where $\theta^{(\l)}_{i}(t)$ is the threshold of the $i^{th}$ neuron in layer $\l$ at time $t$. This threshold is adaptive and unique for each neuron. Only the maximally activated neuron (or \emph{winner}) in a layer is assigned the state 1 if its threshold is exceeded. Our model implements the winner-take-all mechanism which allows only the neuron of highest activity to learn. We say a neuron has \emph{fired} if its state reaches $1$.

Thus, a neuron integrates all inputs over its spatial and temporal RF until it reaches its threshold when it fires if it is the winner. As soon as it fires or if it fails to fire within the duration of its temporal RF, it discharges and then starts integrating again. The discharge from a neuron inhibits neighboring neurons in its own layer. As in \cite{Einhauseretal2002}, it may be assumed that this lateral inhibition is proportional to a neuron's total accumulated charge (or activation) and operates at a faster time scale. The inhibition is required to ensure that neurons in a layer do not get tuned to the same feature set. The inhibition influences a neuron's activation which in turn influences its inhibition. This cycle ensues until a stable state is reached. In most practical cases, this inhibition is observed to be strong enough to drive all neurons close to their baseline activation. In our implementation, we assume this baseline to be zero which does not effect our features qualitatively.

\subsection{Updating weights and thresholds}
\label{Sec:Updating weights and thresholds}

Feedforward weights to neuron $j$ in layer $\l$ with $S^{(\l)}_{j}(t)=1$ are updated following Hebbian rule.

\begin{equation}
\label{Equ:Bottom-up weight update}
W^{(\l-1,\l)}_{ij}(t+1) = (1-\alpha) \times W^{(\l-1,\l)}_{ij}(t) + \alpha \times S^{(\l-1)}_{i}(h)
\end{equation}

\noindent where $t_{0}\leq h\leq t$, $\alpha$ is the learning rate that decreases with time for finer convergence, $0<\alpha <1$, $S^{(0)}=A^{(0)}$. This weight update rule is obtained by applying gradient descent on the objective function in equ. \ref{Equ:Objective function} in an online setting. Feedforward weights leading to each neuron are initialized to ones and normalized to have unit norm, which allows all neurons in a layer to compete on an equal footing. A new neuron is not recruited unless the incoming pattern is more similar to the initialized feature than to any of the learned features. After each update, weights to each neuron are normalized to have unit norm. Thus, feedforward connection from a presynaptic neuron ($i$) to a postsynaptic one ($j$) that fire together are strengthened while the rest (to $j$) are weakened. The weakening of connections is crucial for robustness as it helps remove infrequent coincident patterns from memory which are probably noise.

In $L_{1}$, the lateral weight from neuron $i$ to $j$ is also updated following Hebbian rule as:

\begin{equation}
\label{Equ:Lateral weight update}
W^{(1,1)}_{ij}(t+1) = (1-\alpha) \times W^{(1,1)}_{ij}(t) + \alpha \times S^{(1)}_{i}(t-1) \times S^{(1)}_{j}(t)
\end{equation}

\noindent Thus, connection from a presynaptic neuron ($i$) to a postsynaptic one ($j$) that fire at consecutive time instants are strengthened while the rest (from $i$) are weakened. The weights are randomly initialized in $(0.5-\delta,0.5+\delta)$, $\delta\rightarrow 0$, such that $\sum_{j} W^{(1,1)}_{ij}=1$, and the above learning rule ensures that constraint continues to be satisfied. Since $S^{(1)}$ is extremely sparse, $W^{(1,1)}$ can store a number of patterns from their correlations at consecutive time instants.

The threshold is updated as follows:

\begin{equation}
\label{Equ:Threshold update}
\theta^{(\l)}_{i}(t+1) =
  \left\{
  \begin{array}{ll}
    A^{(\l)}_{i}(t), & \mbox{if $S^{(\l)}_{i}(t)=1$}
\\
    (1-\eta) \times \theta^{(\l)}_{i}(t), & \mbox{if $S^{(\l)}_{i}(t)=0$ and $t-t_{0}=\tau^{(\l)}$}
  \end{array}
  \right.
\end{equation}

\noindent where $\eta$ is the threshold decay parameter, a constant, $0<\eta <1$. Due to the threshold, only a small subset of stimuli can trigger learning. The threshold decay ensures that the size of this subset remains fixed throughout the learning process, thereby maintaining the plasticity of the network. The winner-take-all mechanism along with threshold favor neurons with sparsely distributed activity.

In the proposed model, a winner neuron always passes on its activations to its neighboring neurons in all layers irrespective of whether it fires or not. This is crucial for online operation where learning and inferencing proceed simultaneously and not in distinct phases. If a pattern has been learned and a part of it is shown, a partial pattern of activations will stimulate the remaining neurons of the pattern to become active thereby completing the whole pattern. However, the strength of connections will not be altered unless enough of the pattern has been seen (as determined by $\theta$) and the RFs of the presynaptic neurons are the best match to the incoming pattern to fire the postsynaptic neuron in the higher layer.

\section{Experimental results}
\label{Sec:Experimental results}

The proposed model was deployed for learning visual features in a node from spatiotemporal data in an unsupervised and online manner. The feedforward weights were learned layer by layer with $\alpha(t)=\alpha(t-1)/(1+t/10^{6})$, $\alpha(0)=0.1$. $\theta^{(\l)}$ were initialized to a value slightly greater than $\tau^{(\l)}$ such that the longest sequences may be captured. $\eta=10^{-6}$. As stimuli we used 17 videos recorded at different natural locations with a CCD camera mounted on a cat's head exploring its environment \cite{Betschetal2004}. These videos provided a continuous stream of stimuli similar to what the cat's visual system is naturally exposed to, preserving its temporal structure. The same catcam videos were used in \cite{Einhauseretal2002,Masquelieretal2007} for evaluating models on learning complex cell RF properties. As preprocessing, each frame ($320\times 240$ pixels) was converted to grayscale and convolved with a $3\times 3$ Laplacian of Gaussian kernel followed by rectification to crudely highlight edges, believed to be performed by center-surround cells before the signal reaches V1. Spatiotemporal voxels of size $10\times 10$ pixels spanning over the entire duration of a video were extracted at fixed points from a $9\times 11$ grid, sampled every 25 pixels. These 99 voxels from each video formed our stimuli, leading to a total of about 5.3 million patches from the 17 videos.

\subsection{Simple layer}
\label{Sec:Simple layer}

\begin{figure}[t]
    \centering
        \includegraphics[width=0.99\textwidth]{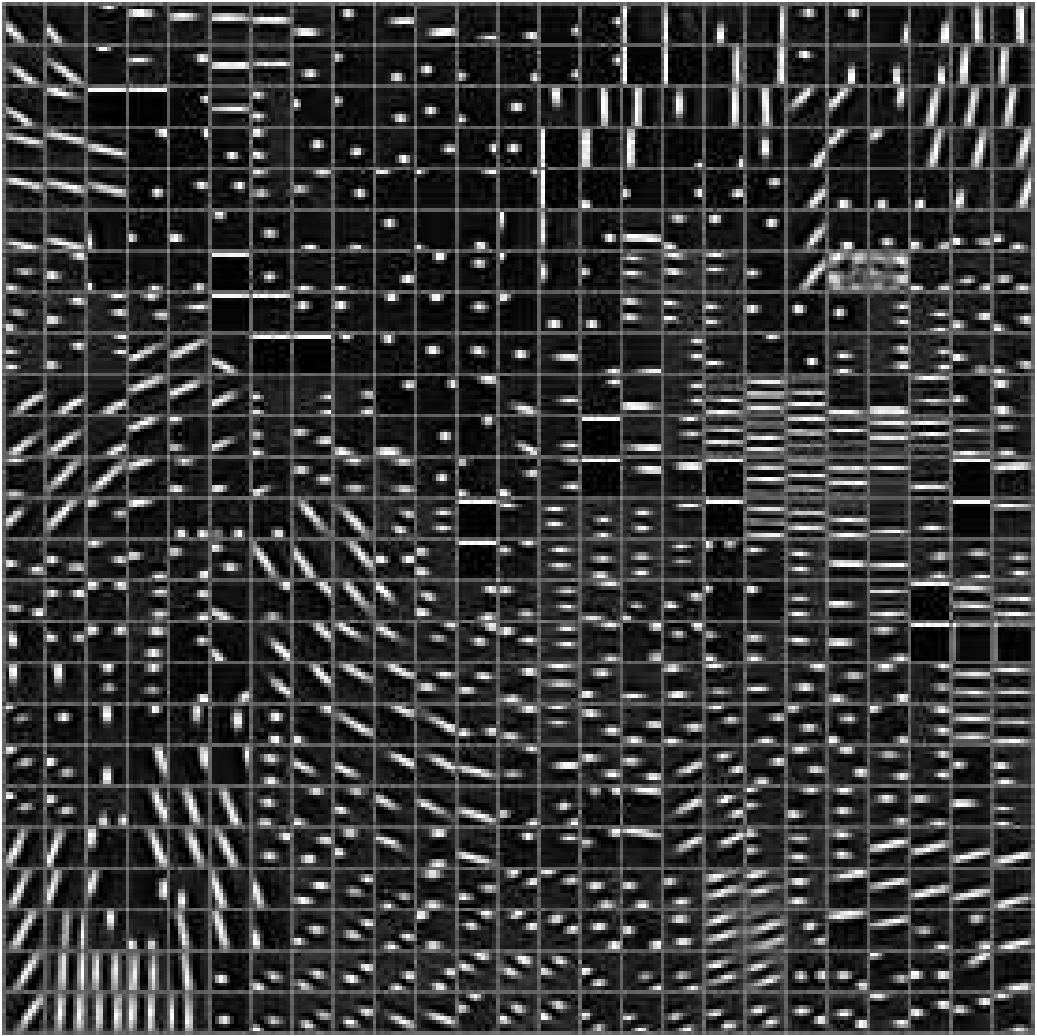}
    \caption{Features learned by 625 neurons in $L_{1}$ from the catcam video.}
\label{Fig:SimpleFeatures} 
\end{figure}

Our model was simulated with 625 simple neurons in $L_{1}$ with spatial RF size $10\times 10$ pixels. Each simple neuron learned a unique visual feature from the stimuli. Qualitatively, the features belonged to three distinct classes of RFs -- small unoriented features, localized and oriented Gabor-like filters, and elongated edge-detectors (see Fig. \ref{Fig:SimpleFeatures}). Such features have been observed in macaque V1, and have been reported to be learned by computational models such as SAILnet \cite{Zylberbergetal2011} and SSC \cite{RehnSommer2007}.

If lateral connections encode transition probabilities and minimization of wiring length is an objective, neurons that fire in close temporal proximity will end up being spatial neighbors. Furthermore, if the stimulus changes gradually, neighboring neurons will develop similar feature preferences. In order to learn features in a topographic map, we organize the simple layer neurons on a 2D grid. At any time $t$, the activation $A_{i}$ of the winner neuron $i$ at time $t-1$ is propagated to its neighbor $j$ ($j\neq i$), the effect of which exponentially decreases with square of the distance $d_{ij}$ between $i$ and $j$ on the grid. For neighbor $j$ at time $t$, the propagated activation is:

\begin{equation}
\psi^{(1)}_{ij}(t) =
  \left\{
  \begin{array}{ll}
    e^{-\gamma\times d_{ij}^{2}}, & \mbox{if $i\neq j$}
\\
    0, & \mbox{otherwise}
  \end{array}
  \right.
\end{equation}

\noindent where $\gamma$ is a constant, $\gamma=2$. At any time $t$, in addition to feedforward activation, each simple neuron receives an activation from a neighboring winner in the same layer. The simple layer activation is:

\begin{equation}
\label{Equ:Topographic activation}
A^{(1)}(t) = A^{(0)}(t)\times W^{(0,1)}(t) + S^{(1)}(t-1)\times\psi^{(1)}(t)
\end{equation}

where $\psi^{(1)} = [\psi^{(1)}_{ij}]_{|L_{1}|\times |L_{1}|}$, $|L_{1}|$ is the number of neurons in $L_{1}$. The second term biases neighboring neurons to become the winner at the next instant. As a result, simple neurons that fire in close temporal proximity end up being spatially close in the 2D grid (hence, equations \ref{Equ:Topographic activation} and \ref{Equ:SimpleActivation} are functionally equivalent). Consequently, the wiring length for pooling by complex neurons is reduced, in agreement with biological evidence \cite{Blasdel1992,DeAngelisetal1999}. The topographic map is shown in Fig. \ref{Fig:SimpleFeatures}. The pooling region in this topographic map as learned by each complex neuron is shown in Fig. \ref{Fig:ComplexFeatures}.

\subsection{Complex layer}
\label{Sec:Complex layer}

Our model was simulated with 25 complex neurons in $L_{2}$ with temporal RF size of 21 sampling instants. Being exposed to the catcam videos, each complex neuron got strongly connected to a subset of simple neurons in $L_{1}$ i.e., it learned a unique transformation to which it is now invariant. The spatial feature encoded by each simple neuron in this subset is an instance of the transformation. The activation of a complex neuron is high if the spatial stimulus matches any of these spatial features, and low otherwise. Thus, the response of complex neurons in our model is akin to that of complex cells in V1.

Due to the nature of stimulus, our model was exposed to sequences of spatial stimuli in the catcam video. Repeating sequences, if learned, would be useful for prediction. When trained with a sequence (e.g., $\langle A,B,C,D,E\rangle$), a complex neuron in our model responds much more vigorously (as measured by its activation) to the corresponding set (e.g., $\{A,B,C,D,E\}$) than to any other (e.g., $\{I,J,K\}$), where each alphabet refers to a unique spatial feature. Further, it responds more vigorously to the training sequence than to any other (e.g., $\langle E,D,C,B,A\rangle$), thereby manifesting the complex neuron's direction selectivity. This is achieved by exploiting the set learned by the complex neuron in conjunction with the transition probabilities learned by the lateral connections in the simple layer. The difference in activations towards the training sequence and any of its other permutation depends on how often other permutations of the set are presented. If no other permutation is presented, the difference in activations is high. In V1, 10-20\% cells show marked direction selectivity \cite{Hubel1995}.

\begin{figure}[htbp]
\centering
\includegraphics[width=0.6\textwidth]{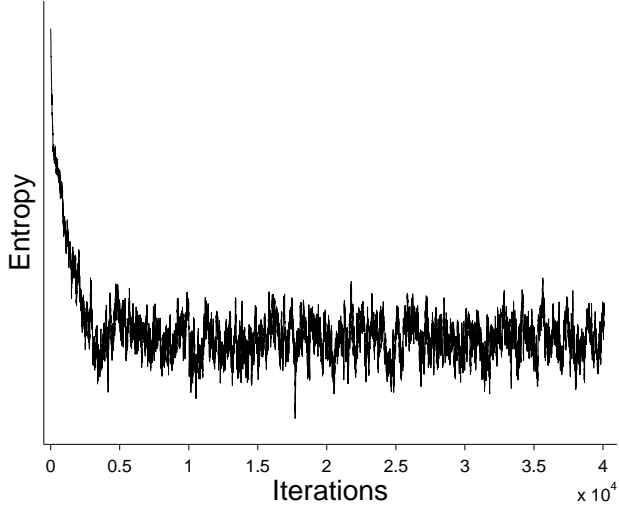}
\caption{Entropy of the system as it learns from natural stimuli.}
\label{Fig:Entropy} 
\end{figure}

\begin{figure*}[t]
\centering
    \subfigure[]
    {
        \label{Fig:L2neuron3}
        \begin{minipage}{0.23\textwidth}
            \begin{center}
            \includegraphics[height=.016\textheight,keepaspectratio]{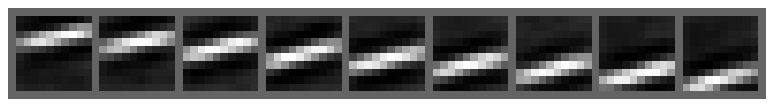}
            \includegraphics[width=\textwidth]{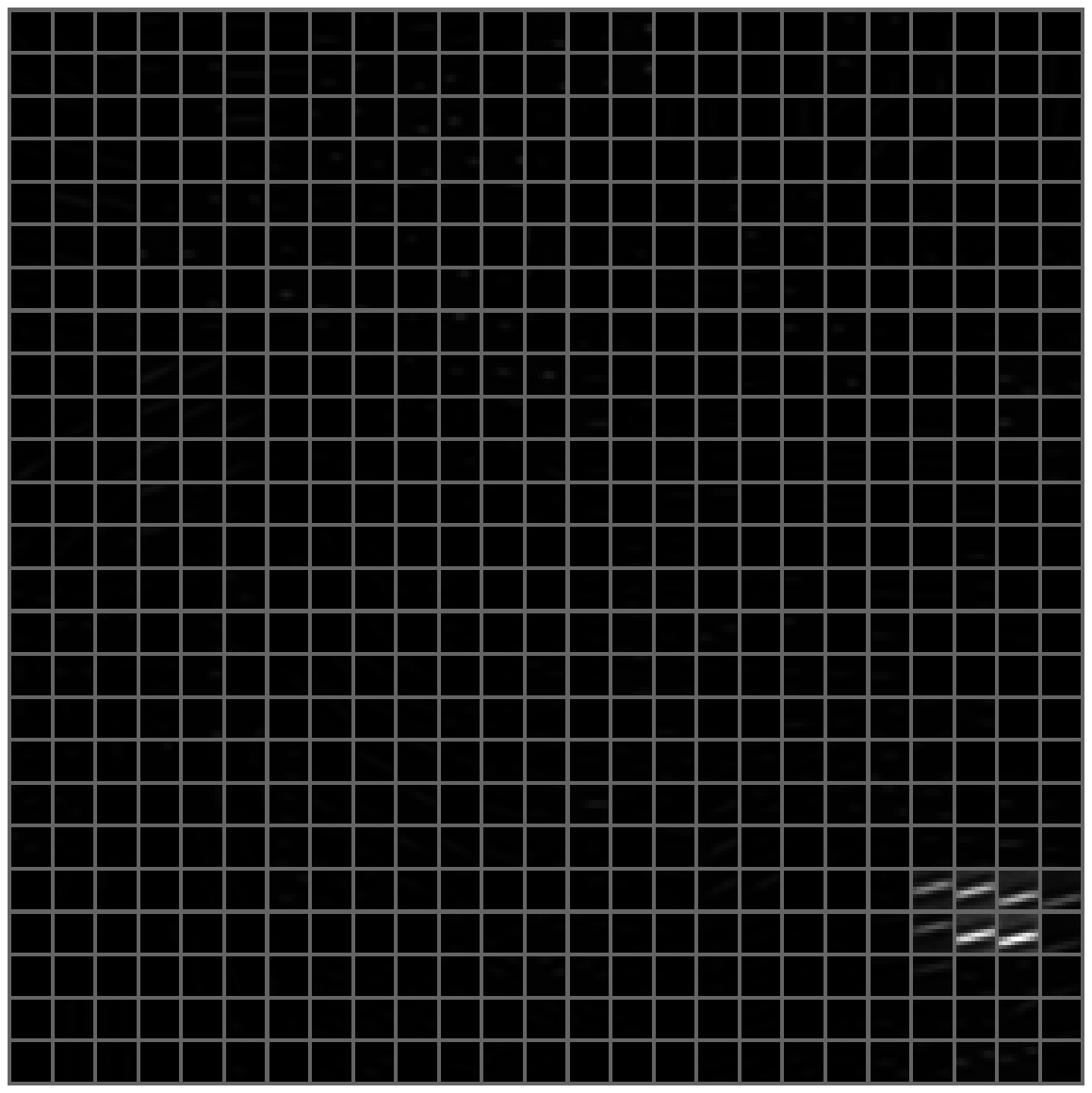}
            \end{center}
        \end{minipage}
    }
    \subfigure[]
    {
        \label{Fig:L2neuron4}
        \begin{minipage}{0.23\textwidth}
            \begin{center}
            \includegraphics[height=.016\textheight,keepaspectratio]{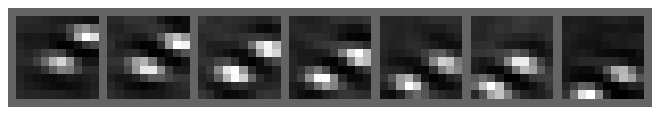}
            \includegraphics[width=\textwidth]{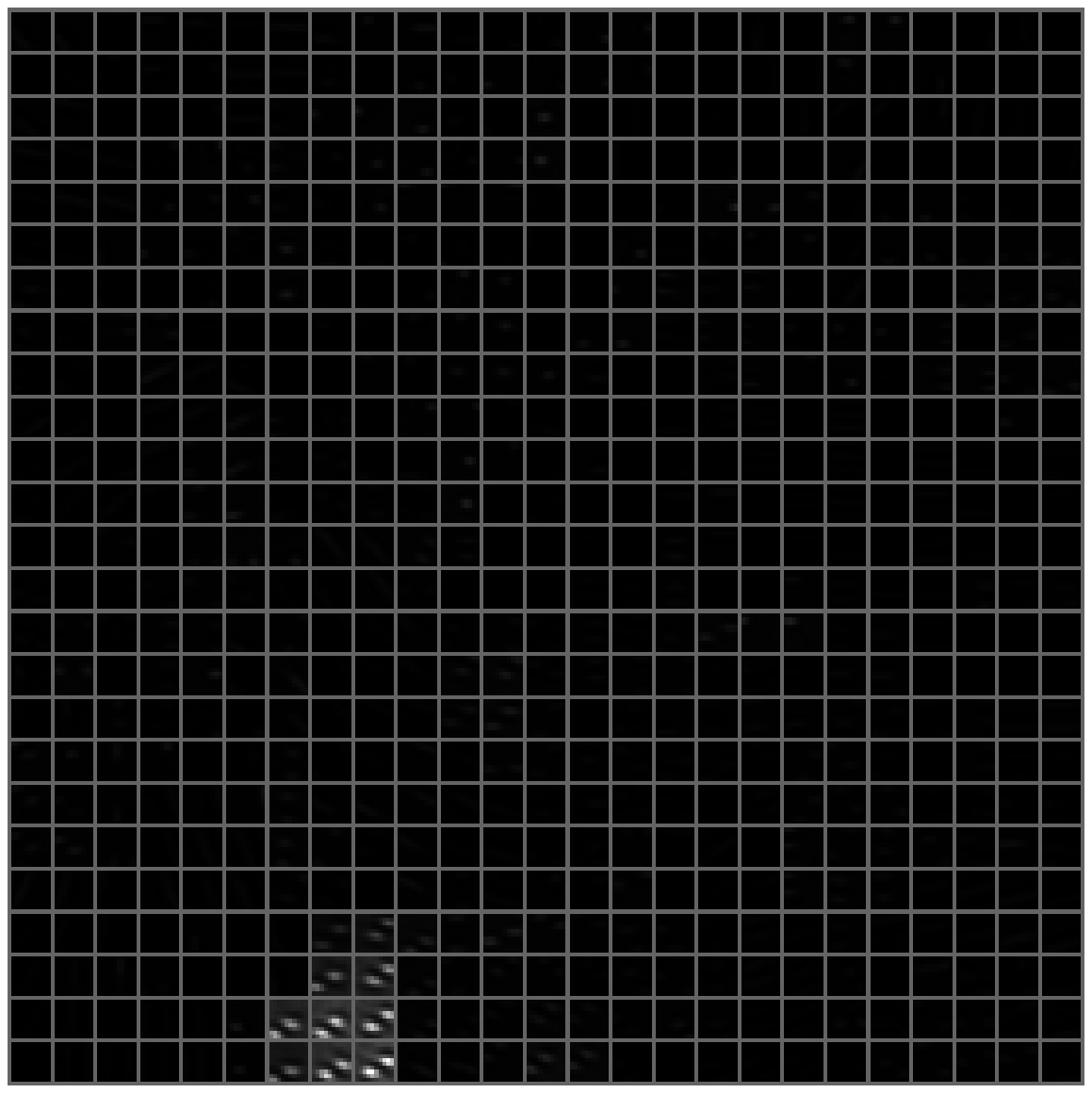}
            \end{center}
        \end{minipage}
    }
    \subfigure[]
    {
        \label{Fig:L2neuron5}
        \begin{minipage}{0.23\textwidth}
            \begin{center}
            \includegraphics[height=.016\textheight,keepaspectratio]{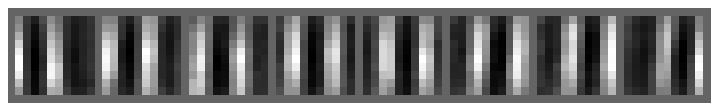}
            \includegraphics[width=\textwidth]{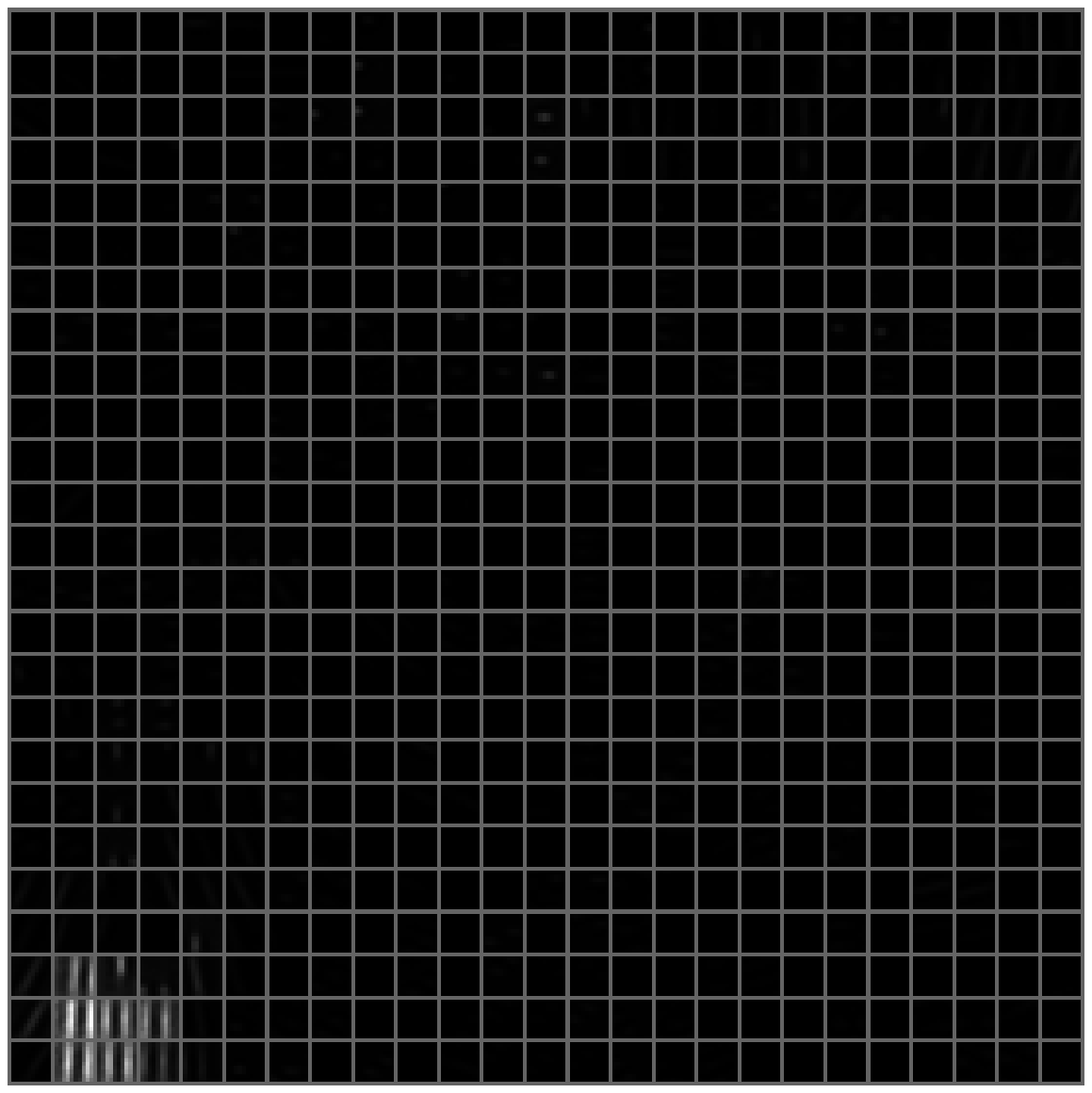}
            \end{center}
        \end{minipage}
    }
    \subfigure[]
    {
        \label{Fig:L2neuron6}
        \begin{minipage}{0.23\textwidth}
            \begin{center}
            \includegraphics[height=.016\textheight,keepaspectratio]{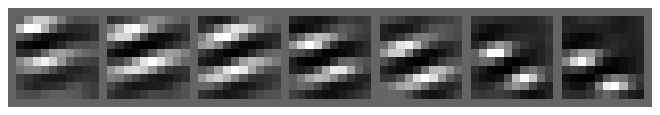}
            \includegraphics[width=\textwidth]{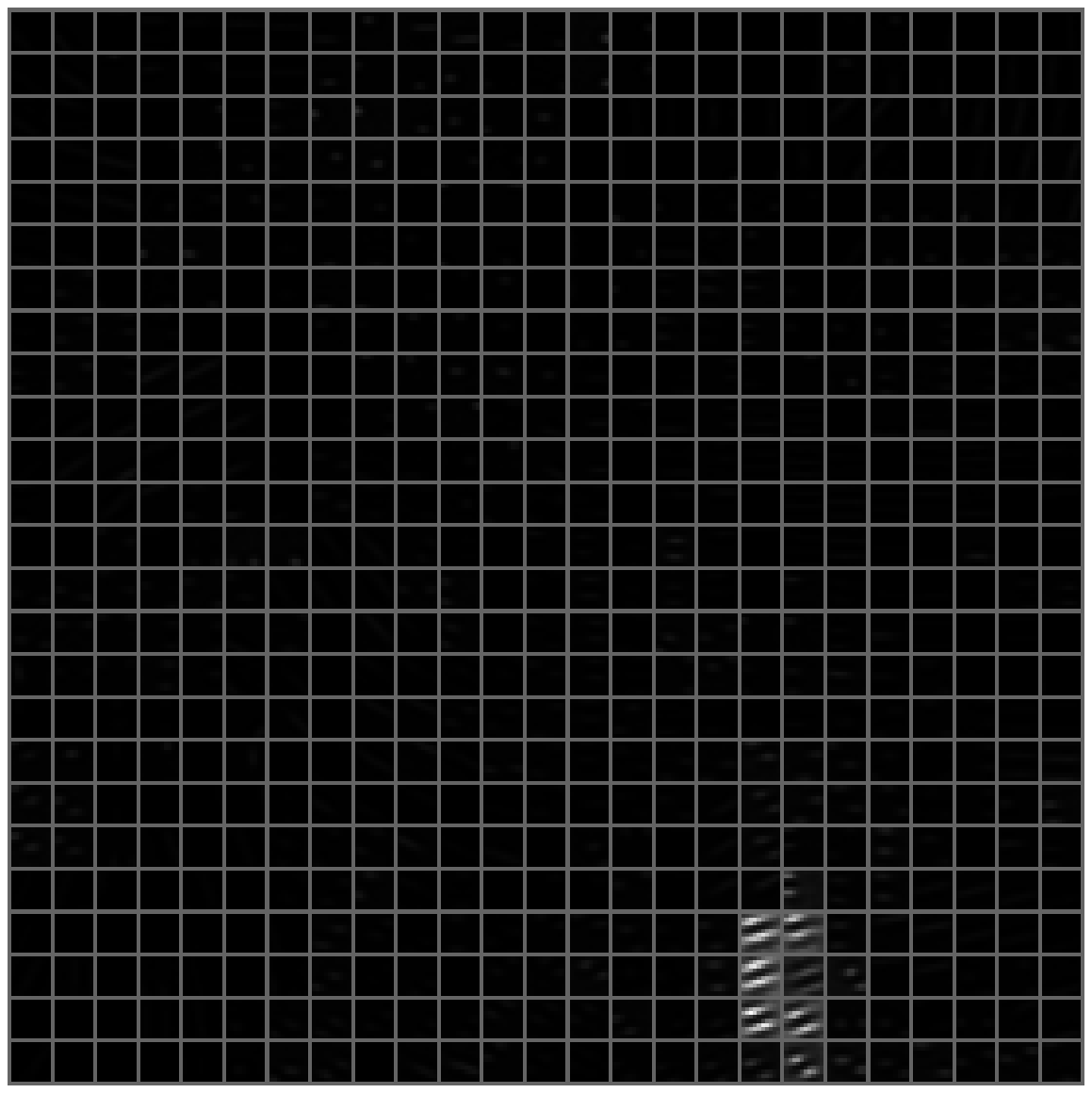}
            \end{center}
        \end{minipage}
    }
    \subfigure[]
    {
        \label{Fig:L2neuron7}
        \begin{minipage}{0.23\textwidth}
            \begin{center}
            \includegraphics[height=.016\textheight,keepaspectratio]{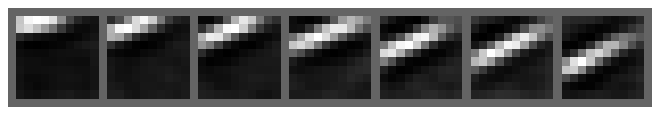}
            \includegraphics[width=\textwidth]{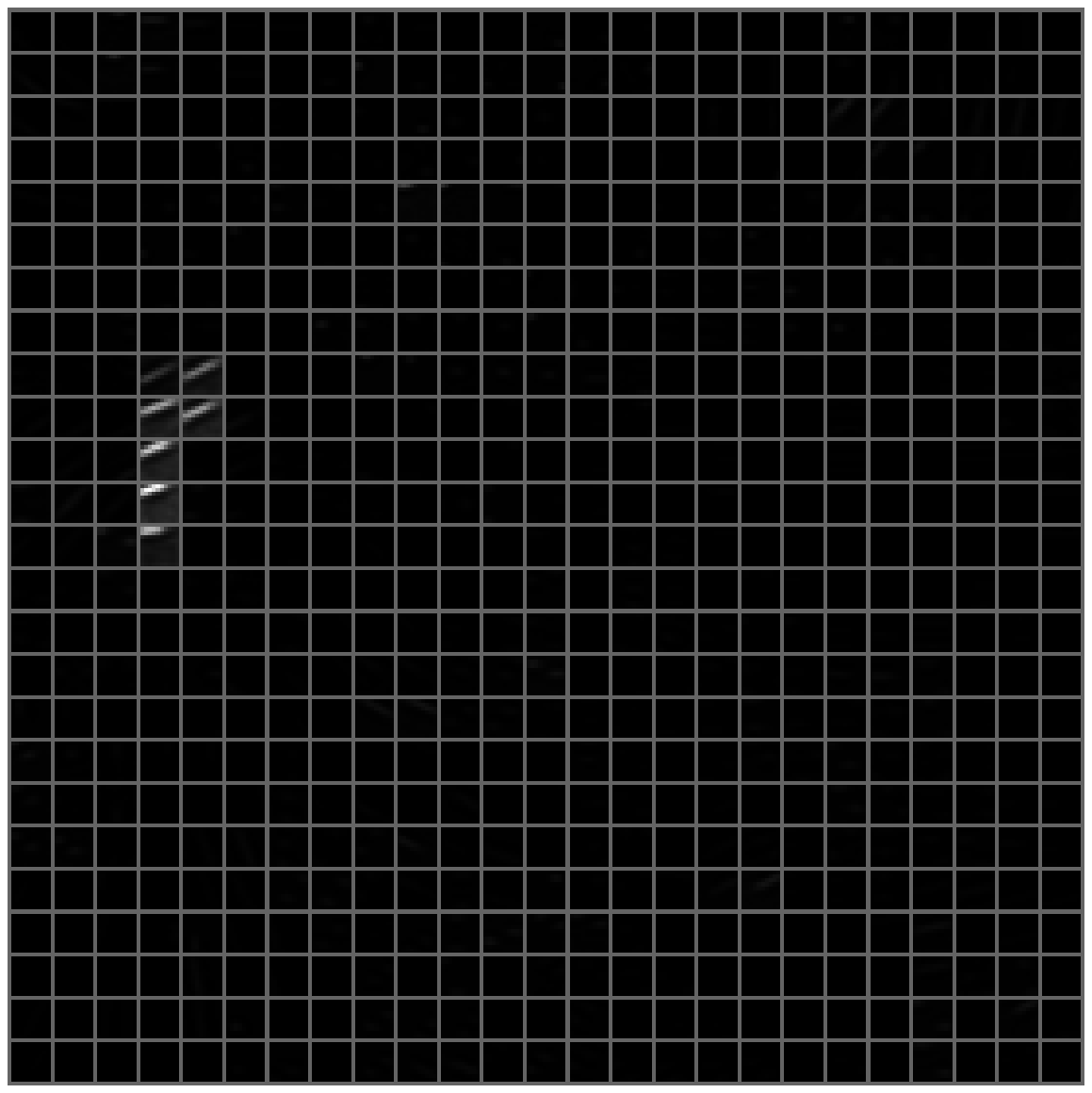}
            \end{center}
        \end{minipage}
    }
    \subfigure[]
    {
        \label{Fig:L2neuron15}
        \begin{minipage}{0.23\textwidth}
            \begin{center}
            \includegraphics[height=.016\textheight,keepaspectratio]{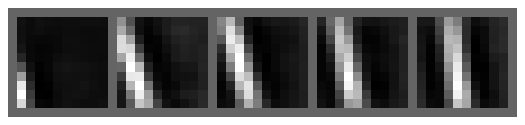}
            \includegraphics[width=\textwidth]{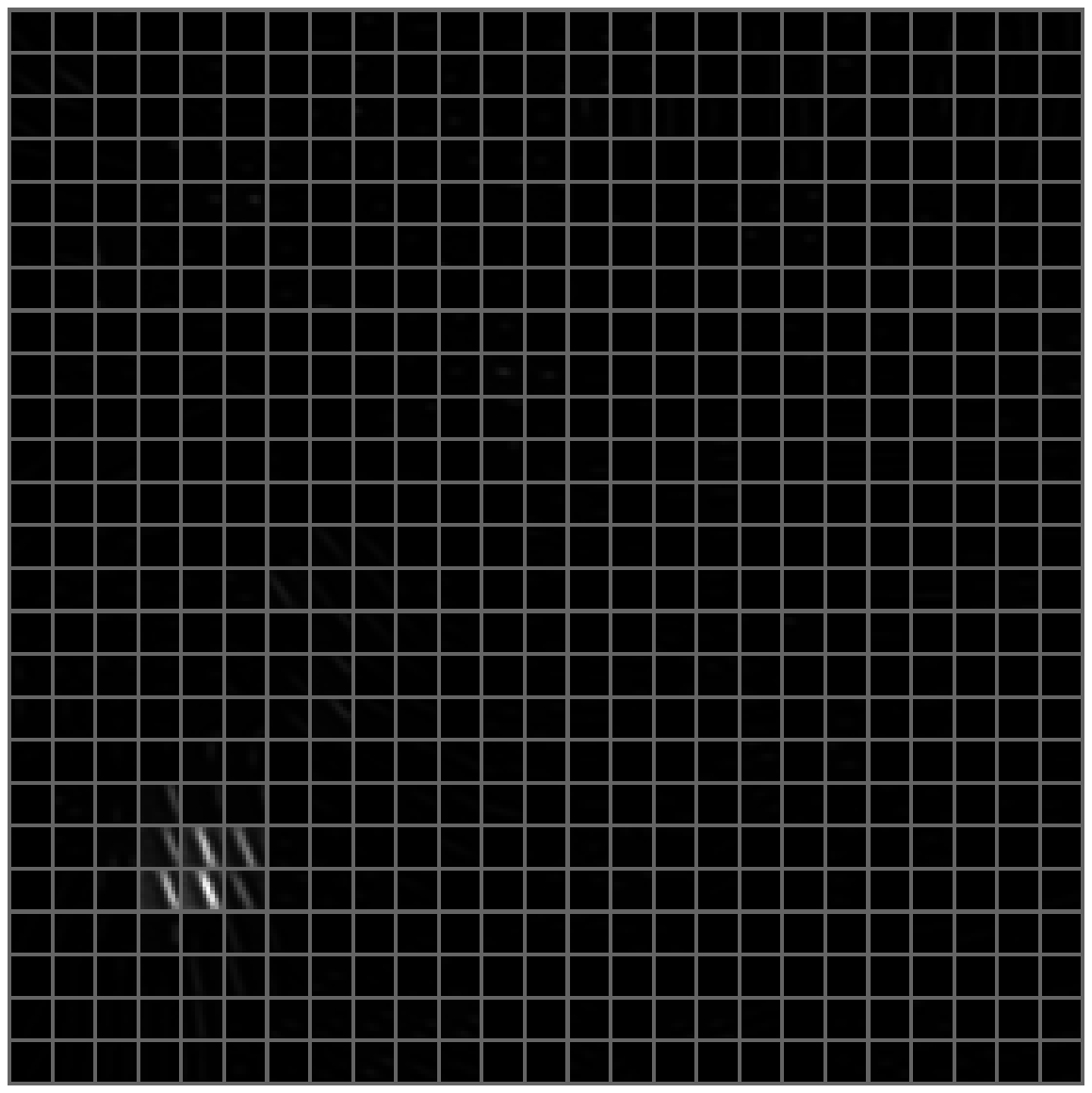}
            \end{center}
        \end{minipage}
    }
    \subfigure[]
    {
        \label{Fig:L2neuron7}
        \begin{minipage}{0.23\textwidth}
            \begin{center}
            \includegraphics[height=.016\textheight,keepaspectratio]{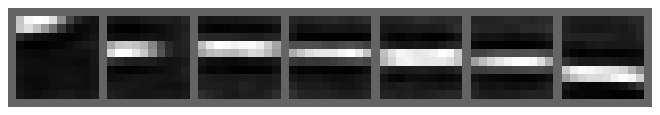}
            \includegraphics[width=\textwidth]{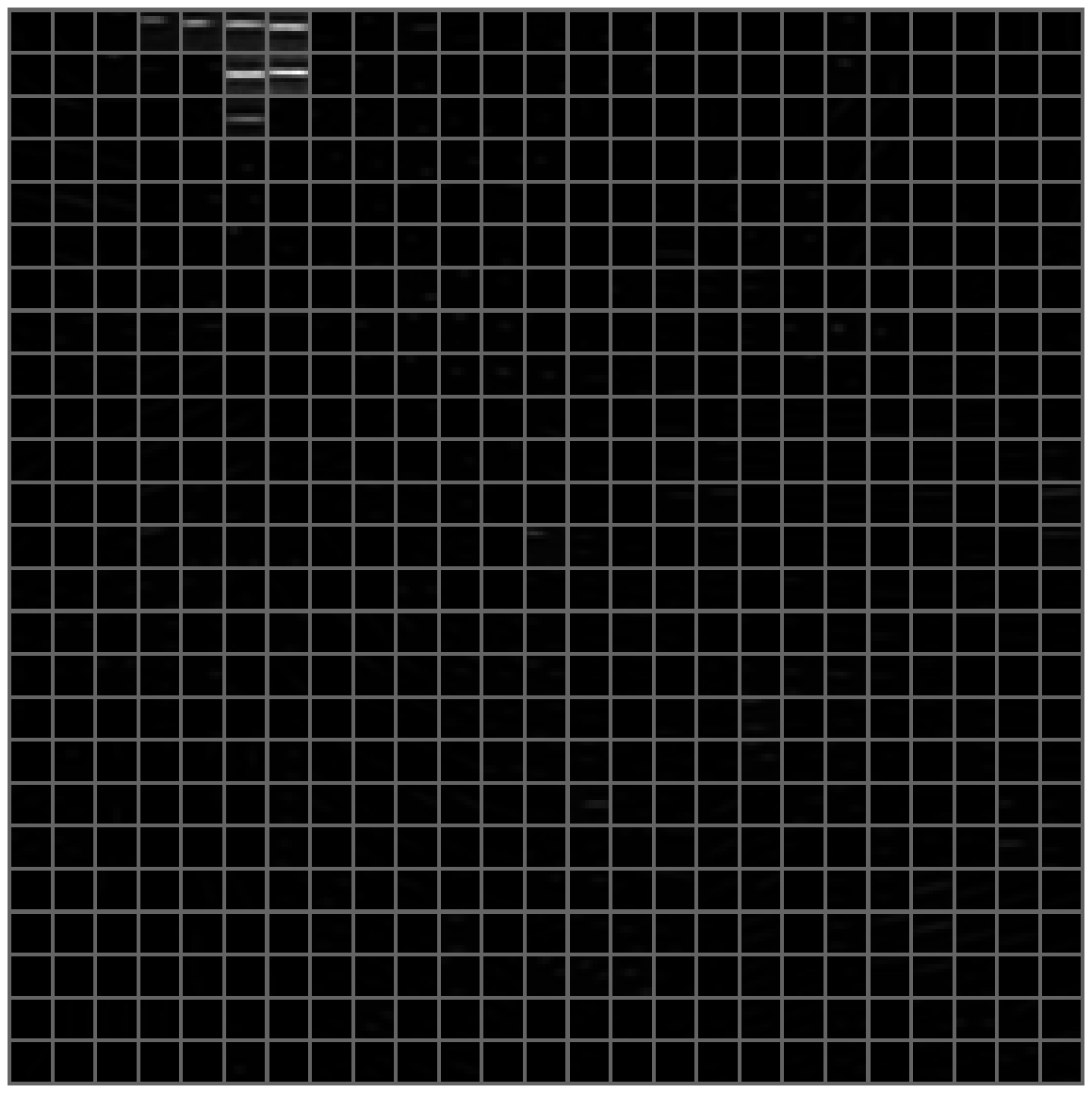}
            \end{center}
        \end{minipage}
    }
    \subfigure[]
    {
        \label{Fig:L2neuron15}
        \begin{minipage}{0.23\textwidth}
            \begin{center}
            \includegraphics[height=.016\textheight,keepaspectratio]{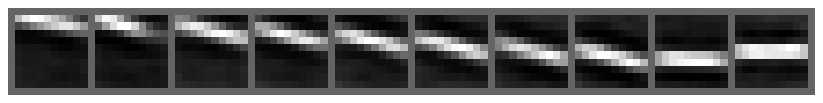}
            \includegraphics[width=\textwidth]{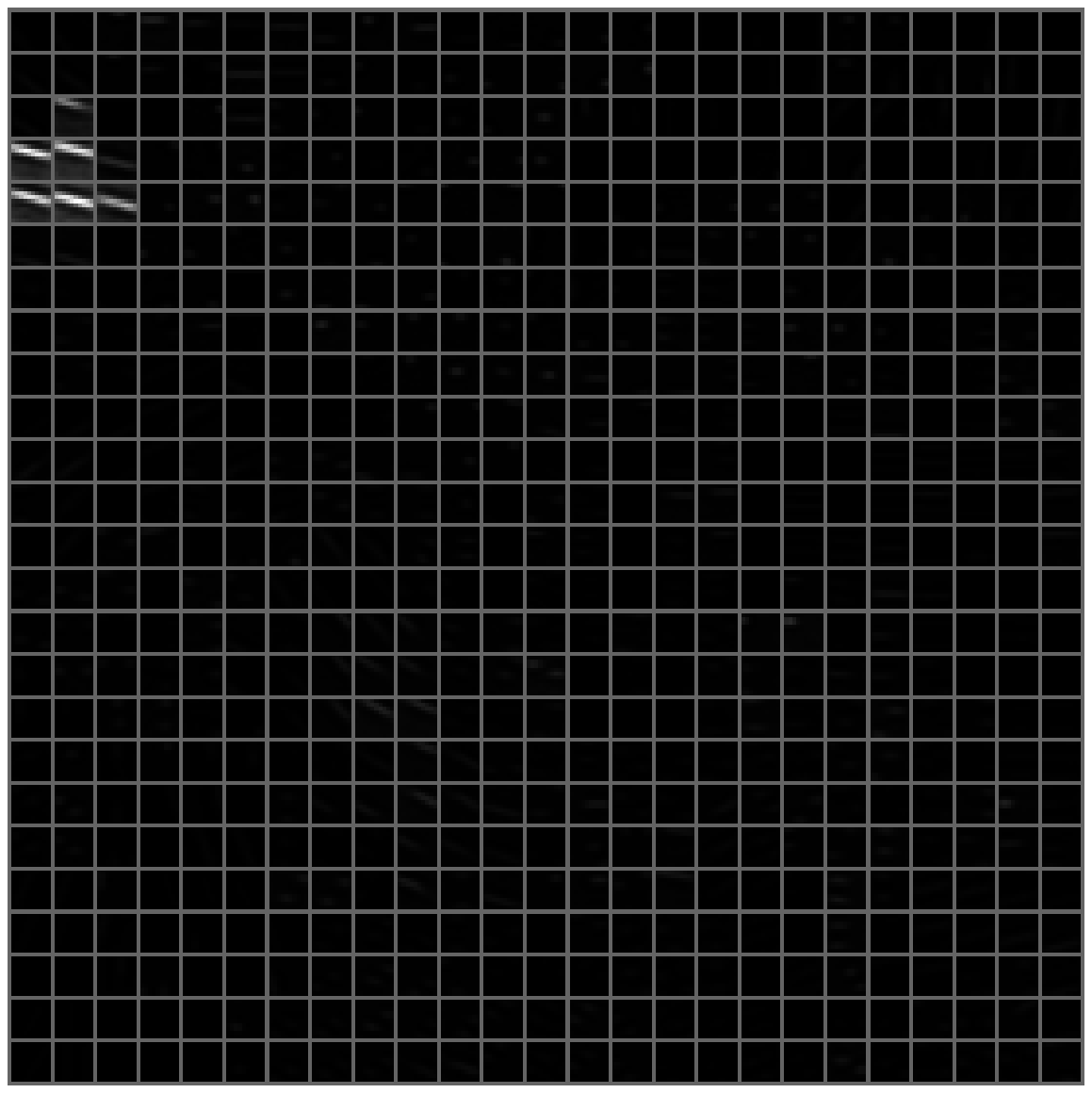}
            \end{center}
        \end{minipage}
    }
    \caption{Sequences and feedforward connection strengths learned by eight (out of 25) $L_{2}$ neurons from the catcam videos are shown in (a) through (h). In (a), the top figure shows the sequence of length 9 learned by this $L_{2}$ neuron. The bottom figure shows the connection strengths to the 625 $L_{1}$ neurons learned by this $L_{2}$ neuron. Similarly for (b) through (h). The $L_{2}$ neurons learn variable length sequences even with the same $\tau^{(2)}$ (=21).}
\label{Fig:ComplexFeatures} 
\end{figure*}

Prediction in our model amounts to computing the probability of the $i^{th}$ simple neuron being the winner at time $t+1$ given that the $j^{th}$ simple neuron was the winner at time $t$, i.e. probability of $S^{(1)}_{i}(t+1)=1$ given $S^{(1)}_{j}(t)=1$, which depends on the transition probabilities as well as the sets learned by the complex neurons. At any instant, the winner complex neuron (say, $k$) restricts the set for the expected winner simple neuron. The highest expected one is then chosen from this set using the transition probabilities.

\begin{equation}
\label{Equ:Prediction}
Pr(S^{(1)}_{i}(t+1)=1~|~S^{(1)}_{j}(t)=1) = \kappa\times (\frac{W^{(1,2)}_{jk}}{\sum_{k}W^{(1,2)}_{jk}}\times\frac{W^{(1,2)}_{ik}}{\sum_{i}W^{(1,2)}_{ik}} + W^{(1,1)}_{ji})
\end{equation}

\noindent where $\kappa$ is the uniform prior distribution. Fig. \ref{Fig:Entropy} shows the entropy of the system as it converges with learning. Fig. \ref{Fig:ComplexFeatures} shows the sets and sequences learned by eight $L_{2}$ neurons in our model. To reconstruct the sequence learned by a $L_{2}$ neuron, we select the strongest connected feature from its set; its successor is that feature from the set that has the strongest lateral connection (the algorithmic implementation of equ. \ref{Equ:Prediction}), and so on until a feature is repeated, signifying the end of sequence.

\section{Conclusions}
\label{Sec:Conclusions}

Learning features invariant to arbitrary transformations in the data is a requirement for any recognition system, biological or artificial. Biological evidence and computational models have supported the role of simple-complex layers in V1 in achieving this goal. To understand their function as a canonical computational unit in a hierarchical or deep network, we presented a novel two-layered neural model that operates in a feedforward, unsupervised and online manner. When exposed to natural videos recorded with a camera mounted on a cat's head, the first layer neurons learned three classes of spatial features that resemble the RFs in macaque V1 while the second layer neurons learned arbitrary transformations in the data, their activations were then invariant to these transformations akin to the response of complex cells in V1. The learning rules for the two layers were derived from the same objective function signifying their functional similarity. The simple and complex RFs were learned by spherical clustering in space and time respectively where the outliers were not allowed to influence the cluster centers.

The model could make higher-order predictions by simultaneously exploiting the transformations learned in the complex layer and transition probabilities learned by the lateral connections in the simple layer. We showed the convergence of this predictive model while learning from the catcam videos. Unlike other models with predefined pooling regions or presumed group sparsity for learning topographic maps from spatial data, we used temporal continuity of data and physical constraints to learn topographic feature map. The proposed model is fully-learnable with only two manually tunable parameters -- the learning rate and threshold decay parameter. We conclude that the model is an ideal candidate to be used as a canonical computational unit in a hierarchical network for real world applications and understanding biological brain functions.

\subsubsection*{Acknowledgments}
Research reported in this paper was partially supported by the U.S. National Science Foundation under CISE Grant No. 1231620.

\bibliographystyle{unsrt}
\bibliography{mybibfile}

\begin{thebibliography}{10}

\bibitem{HubelWiesel1962}
D.~H. Hubel and T.~N. Wiesel.
\newblock Receptive fields, binocular interaction and functional architecture
  in the cat's visual cortex.
\newblock {\em Journal of Physiology}, 160:106--154, 1962.

\bibitem{Fukushima2003}
K.~Fukushima.
\newblock Neocognitron for handwritten digit recognition.
\newblock {\em Neurocomputing}, 51(1):161--180, 2003.

\bibitem{LeCunBengio1995}
Y.~LeCun and Y.~Bengio.
\newblock Convolutional networks for images, speech and time series.
\newblock In M.~A. Arbib, editor, {\em The Handbook of Brain Theory and Neural
  Networks}, pages 255--258. MIT Press, 1995.

\bibitem{Ciresanetal2012}
D.~C. Ciresan, U.~Meier, and J.~Schmidhuber.
\newblock Multi-column deep neural networks for image classification.
\newblock In {\em Proc. Intl. Conf. Computer Vision and Pattern Recognition},
  pages 3642--3649, 2012.

\bibitem{RiesenhuberPoggio1999}
M.~Riesenhuber and T.~Poggio.
\newblock Hierarchical models of object recognition in cortex.
\newblock {\em Nature Neuroscience}, 2(11):1019--1025, 1999.

\bibitem{Serreetal2007recognition}
T.~Serre, L.~Wolf, S.~Bileschi, M.~Riesenhuber, and T.~Poggio.
\newblock Robust object recognition with cortex-like mechanisms.
\newblock {\em IEEE Trans. Pattern Analysis and Machine Intelligence},
  29:411--426, 2007.

\bibitem{DhillonModha2001}
I.~S. Dhillon and D.~S. Modha.
\newblock Concept decompositions for large sparse text data using clustering.
\newblock {\em Machine Learning}, 42(1-2):143--175, 2001.

\bibitem{HyvarinenHoyer2001}
A.~Hyvarinen and P.~O. Hoyer.
\newblock A two-layer sparse coding model learns simple and complex cell
  receptive fields and topography from natural images.
\newblock {\em Vision Research}, 41(18):2413--2423, 2001.

\bibitem{Kavukcuogluetal2009}
K.~Kavukcuoglu, M.~A. Ranzato, R.~Fergus, and Y.~LeCun.
\newblock Learning invariant features through topographic filter maps.
\newblock In {\em Proc. Intl. Conf. Computer Vision and Pattern Recognition},
  2009.

\bibitem{BagonGalun2011}
S.~Bagon and M.~Galun.
\newblock Large scale correlation clustering optimization.
\newblock {\em Computing Research Repository}, arXiv:1112.2903, 2011.

\bibitem{Serreetal2007categorization}
T.~Serre, A.~Oliva, and T.~Poggio.
\newblock A feedforward architecture accounts for rapid categorization.
\newblock {\em Proc. Natl. Academy of Sciences}, 104(15):6424--6429, 2007.

\bibitem{Foldiak1990}
P.~F{\"o}ldi{\'a}k.
\newblock Forming sparse representations by local anti-hebbian learning.
\newblock {\em Biological Cybernetics}, 64:165--170, 1990.

\bibitem{Einhauseretal2002}
W.~Einh{\"a}user, C.~Kayser, P.~K{\"o}nig, and K.~P. K{\"o}rding.
\newblock Learning the invariance properties of complex cells from their
  responses to natural stimuli.
\newblock {\em European Journal of Neuroscience}, 15(3):475--486, 2002.

\bibitem{Betschetal2004}
B.~Y. Betsch, W.~Einh\"{a}user, K.~P. K\"{o}rding, and P.~K\"{o}nig.
\newblock The world from a cat's perspective -- statistics of natural videos.
\newblock {\em Biological Cybernetics}, 90(1):41--50, 2004.

\bibitem{Masquelieretal2007}
T.~Masquelier, T.~Serre, S.~Thorpe, and T.~Poggio.
\newblock Learning complex cell invariance from natural videos{: A}
  plausibility proof.
\newblock Technical Report~60, MIT, Cambridge, MA, December 2007.

\bibitem{Zylberbergetal2011}
J.~Zylberberg, J.~T. Murphy, and M.~R. DeWeese.
\newblock A sparse coding model with synaptically local plasticity and spiking
  neurons can account for the diverse shapes of {V1} simple cell receptive
  fields.
\newblock {\em PLoS Computational Biology}, 7(10):e1002250, 2011.

\bibitem{RehnSommer2007}
M.~Rehn and F.~T. Sommer.
\newblock A network that uses few active neurones to code visual input predicts
  the diverse shapes of cortical receptive fields.
\newblock {\em Journal of Computational Neuroscience}, 22(2):135--146, 2007.

\bibitem{Blasdel1992}
G.~G. Blasdel.
\newblock Orientation selectivity, preference, and continuity in monkey striate
  cortex.
\newblock {\em Journal of Neuroscience}, 12(8):3139–--3161, 1992.

\bibitem{DeAngelisetal1999}
G.~C. DeAngelis, G.~M. Ghose, I.~Ohzawa, and R.~D. Freeman.
\newblock Functional micro-organization of primary visual cortex: Receptive
  field analysis of nearby neurons.
\newblock {\em Journal of Neuroscience}, 19(10):4046–--4064, 1999.

\bibitem{Hubel1995}
D.~H. Hubel.
\newblock {\em Eye, Brain, and Vision}.
\newblock W. H. Freeman, 2nd edition, May 1995.

\end{thebibliography}

\end{document}